# From images in the wild to video-informed image classification

Marc Böhlen
Department of ART
Computational Media
University at Buffalo
Buffalo, NY, USA
marcbohlen@protonmail.com

Raunaq Jain
Department of Computer
Science and Engineering
University at Buffalo
Buffalo, NY, USA
raunaqja@buffalo.edu

Wawan Sujarwo
Ethnobiology Research Group
National Research and
Innovation Agency (BRIN)
Cibinong, West Java, Indonesia
wawan.sujarwo@lipi.go.id

Varun Chandola
Department of Computer
Science and Engineering
University at Buffalo
Buffalo, NY, USA
chandola@buffalo.edu

*Abstract*—Image classifiers work effectively when applied on structured images, yet they often fail when applied on images with very high visual complexity. This paper describes experiments applying state-of-the-art object classifiers toward a unique set of 'images in the wild' with high visual complexity collected on the island of Bali. The text describes differences between actual images in the wild and images from Imagenet, and then discusses a novel approach combining informational cues particular to video with an ensemble of imperfect classifiers in order to improve classification results on video sourced images of plants in the wild.

## I. Introduction

Artificial intelligence (AI) systems are being increasingly deployed to monitor biodiversity [8], encourage conservation [3] and support sustainable development goals [16]. This paper describes ongoing research into the application of AI toward the field of ethnobotany, a practice that seeks to understand how humans utilize and cohabitate with plants, a research endeavor that has only recently begun to deploy AI technologies.

## II. Related work

There is a substantial recent body of research regarding the automated identification of species. From a technical standpoint, the research on Convolutional Neural Network (CNN) architectures for species identification is relevant to this project. Of particular interest are: [12], an approach to image based hierarchical plant classification with images in the wild; [13], the 2018 Expert LifeClef approach to automated plant species identification with fine-tuned high dimensional CNNs; and [6][7], an account of the 2019 LifeCLEF plant identification challenge that sought to address classification of data deficient tropical plants [6]. In particular, the plant identification challenges, or PlantCLEF, conducted as part of the LifeCLEF challenge have in recent years reported a marked increase in the accuracy of classification results produced by machine learning methods. While the training data sizes varied over the years, the datasets typically consisted of 100,000 images of plants belonging to approximately 10,000 species. In the 2017 challenge [5], CNN based methods reported a *mean reciprocal rate* (MRR) of over 0.90 on a curated data set, and a rate of around 0.80 MRR on a noisy data set (obtained on images collected from the internet). A similar performance was recorded in an enhanced dataset in the 2018 challenge, in which the performance of the deep learning methods was shown to be very close to that of human experts [7]. Interestingly, when the geographical area for the plant species was changed from Europe to South America in the 2019 challenge, the performance of the strongest deep learning methods dropped sharply to 0.30 MRR [6], highlighting the sensitivity of such methods to the geographical distribution of their underlying species and their associated datasets.

This paper describes a set of imagery collected in the context of applied AI in ethnobotany that challenges the current conception of 'images in the wild' that contain much more information than image data in canonical machine learning datasets such as Imagenet. We apply several state-of-the-art classifiers onto this new image set and demonstrate the limitations of state-of-the-art classifiers in this context. We redirect the difficulties encountered in studying wild images towards reflecting on image collection in general, and propose a video supported -approach to image classification that makes use of video-specific information in support of image classification.

### A. Field study site, data collection and the bali-26 dataset

We have established a field study site in Central Bali, Indonesia. The selection of the field study site is motivated by the fact that Bali is home to luscious and diverse flora and a long history of local ethnobotanical traditions. However, these practices are in decline on the island, specifically amongst younger generations, and some areas of the island are subject to intensifying land use and touristic activities, reducing protected areas for some plants. As such, Bali is a study site that will directly profit from advances in recording technologies and making local ethnobotanical knowledge that might otherwise in fact be lost amendable to machine learning.

Our process begins with the compilation of a list of significant plants in consultation with an expert in Balinese ethnobotany. This list is passed on to the project PI and data collectors in the field. The team discusses the individual plants [2] and suggests possible field sites. The data collectors then individually collect mobile phone-sourced video footage of the plants in the wild, and send this footage to the research team. Data collectors are then immediately remunerated by electronic payment. We then extract individual images from these video feeds in a custom-built procedure, coined *Catch&Release*[1], which allows us to label the images either with the information contained in the audio track or by manually setting the labels.

---

[1] https://github.com/realtechsupport/c-plus-r

*The bali-26 collection*[2] currently includes 26 distinct plant species of particular significance to Balinese ethnobotany, collected through the procedures described above. Bali-26 mainly contains plant species from the Malesian floristics region (Southeast Asian countries), with several of these species having their origins in Bali. The bali-26 collection is to our knowledge the first attempt to make the rich ethnobotanical legacy and knowledge of Southeast Asia and Bali, in particular, accessible to machine learning procedures.

All plant species contained in the bali-26 collection are represented with at least 1200 images, and in some cases over 2000 instances. In each case we represent the species with images of leaves, fruit, bark or stem in various stages of growth. As such, our collection is under-specified; each of the 26 categories contains multiple facets of the species, including views from various distances and shots of the plants readied for consumption at markets. We lump all of these discrete instances into a single representative plant label. This configuration follows directly from the field data collection approach. However, the approach makes classification much more challenging as the image material contains more varied information than botany specific plant databases. The differences are illustrated in figures 1 and 2, each showing a representation of plants readied for classification as represented by the *Swedish Leaf Dataset*[3] and the bali-26 dataset.

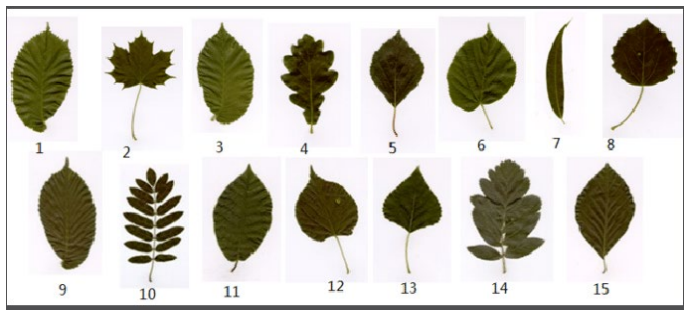

Fig. 1.    Samples contained in the Swedish Leaf Dataset

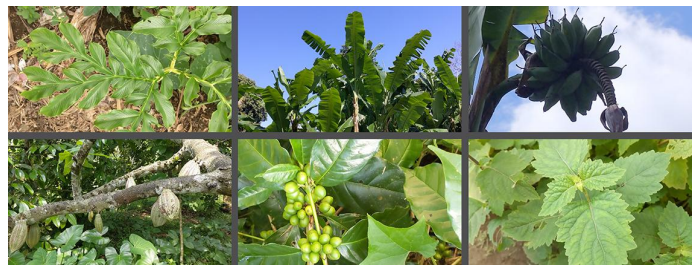

Fig. 2.    Samples from the bali-26 dataset. Top left: Elephant foot yam (leaves); top center: Banana (leaves), top right: Banana (fruits on tree). Bottom left: Cacao (fruits on tree), bottom center Coffee Arabica (leaves and fruit, not ripe); bottom right: Patchouli (leaves). All images extracted from videos captured in the field.

### III.    IMAGE COMPLEXITY

We calculated entropy across the images and in local contexts contained in 1022 instances of the bali-26 dataset and in 1000 representative *ImageNet* samples.

The Shannon entropy is almost 0.5 bit larger in the bali-26 than in Imagenet. Local entropy (10 pixels) in both full images and subsampled images is on average over 0.5 bit larger in bali-26 than in Imagenet. The bali-26 images are substantially more complex than the Imagenet images as illustrated in figure 3.

TABLE 1. Summary of information contained within the subsampled images ingested by the CNN architectures

|  | Imagenet samples | bali-26 samples | bali-26 samples |
|---|---|---|---|
| number of images | 1000 | 1022 | 1022 |
| image size | 300 x 300 | 300 x 300 | 1080 x 1920 |
| mean Shannon entropy green | 7.050 | 7.447 | 7.466 |
| mean Shannon entropy red | **7.043** | **7.436** | **7.455** |
| mean Shannon entropy blue | 6.922 | 7.281 | 7.262 |
| mean local 10px entropy green | **3.819** | **4.585** | - |
| std local 10px entropy green | 1.118 | 0.745 | - |
| max local 10px entropy green | 5.997 | 6.0 | - |
| min local 10px entropy green | 0.620 | 1.040 | - |

We performed a principal component analysis (PCA) on the two datasets and calculated the cumulative explained variance for images resized to 200x200 pixels and reduced to grayscale. The top 3 components of the Imagenet dataset capture 48% of the total variance, while the top 3 components in the bali-26 dataset *capture only 20% of the total variance*. The Imagenet data set is easier for PCA to process than the bali-26 data. With these insights, we attempted to identify the most appropriate classification approach to cope with the complexity in our images.

### IV.    CLASSIFICATION APPROACHES

Ethnobotany does not study plant life in isolation, but rather in the context of its environment. Part of this contextualization includes the observation of plants in the wild, and the wild never really contains isolated plants but a jumbled assembly of different plants in different stages of development and decay. From the perspective of image analysis, opening the data collection to unstructured environments substantially complicates the task, as the small study above demonstrates. In order to understand this problem more clearly and to evaluate how state-of-the-art image classifiers respond to our data, we fine-tuned state-of-the-art object detection models on a hand-labelled dataset. While our images had already been labeled with the *Catch&Release* image labelling approach, we applied bounding boxes to a random subset (94 instances of 26 categories) of these images, indicating in all cases within the labeled images specific areas of interest with bounding boxes with the help of *LabelImg*[4], an open-source graphical box annotation tool. After this box labeling process, we applied our data toward state-of-the-art object detection algorithms as described below. All our models were fine-tuned on this small but precise manually generated bounding box label set, creating a uniform baseline for all subsequent experiments and evaluations.

---

[2] https://filedn.com/lqzjnYhpY3yQ7BdfTulG1yY/bali-26.zip
[3] https://www.cvl.isy.liu.se/en/research/datasets/swedish-leaf/
[4] https://github.com/tzutalin/labelImg

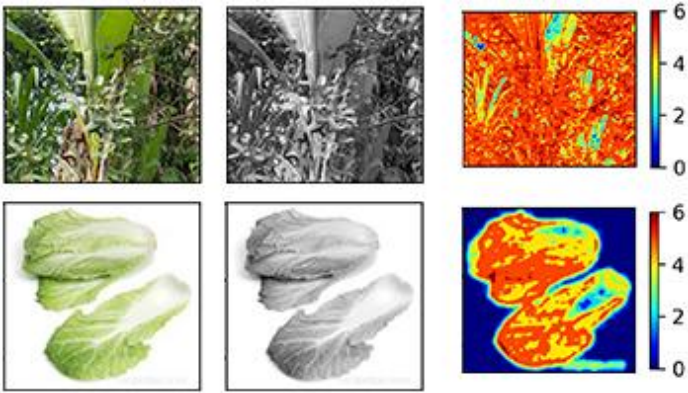

Fig. 3. Visualization of the local entropy in typical images from the bali-26 and Imagenet datasets. Top: Bali-26 mixed plants in the wild (bamboo, banana, elephant grass, guava). Bottom: Imagenet cabbage. Right column: local entropy.

### A. Object detection with region proposal

Object detection approaches that utilize region proposal methods deliver state-of-the-art results [9] for semantic segmentation. Region proposal Convolution Neural Networks, R-CNN, are CNNs with embedded region proposal networks designed to surmise object locations within a given input image. A region proposal network is a fully convolutional network that both predicts object bounds and object scores at each position. While the region proposal increases the ability of a CNN to detect specific objects, it also increases the computational cost of the overall calculation. The Faster R-CNN approach extends the R-CNN approach to exploit full-image convolutional features with the detection network, significantly reducing the computational cost of generating region proposals while retaining the advantage of the subnetwork that tells the unified network "where to look" [10]. The *Detectron2* [17] implementation of the Faster R-CNN model used in our experiments offers several pretrained backbone combinations based on ResNet, ResNext and feature pyramid networks (FPN) for detection at all scales. Specifically, we evaluate four different Detectron2 variants as described below.

TABLE 2. Summary of Detectron2 variants used in the experiments

| Model | Configuration |
|---|---|
| D2.A | Faster R-CNN with Resnet depth of 101 using single feature map at C4 spatial scale - faster_rcnn_R_101_C4_3x |
| D2.B | Faster R-CNN with Resnet depth of 101 using feature pyramid networks (FPN) - faster_rcnn_R_101_FPN_3x |
| D2.C | Faster R-CNN with Resnet depth of 50 using single feature map at C4 spatial scale - faster_rcnn_R_50_C4_1x |
| D2.D | Faster R-CNN with ResneXt backbone - faster_rcnn_X_101_32x8d_FPN_3 |

### B. Bali-26 dataset - Multi-instance classification with a single classifier for individual plants

In this experiment, we seek to detect multiple instances of a single plant in one image. This problem extends prior work [2] on classifying images of plant instances from video. The first evaluation set, bali-26 samples, contains 1000 samples from the 26 categories (40 randomly sampled images per category). Each image contains one plant species as shown in figure 4.

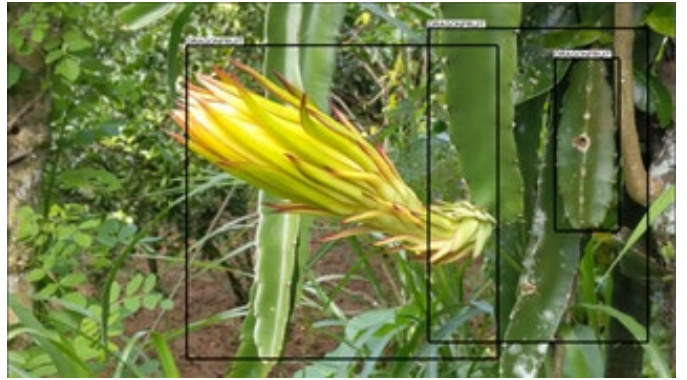

Fig. 4. Detectron2 D2.A result on bali-16 dataset. Classifier detecting dragonfruit, both mature flower (left box) and leaves (right box)

### C. Synthetic image set - multi-instance classification with a single classifier on composite plant images

In this experiment, we created composite images from the bali-26 dataset with two different plant species (top and bottom or left and right). This image set serves as an intermediate complexity test that contains some challenging features of images in the wild while reducing the 'wild' to a manually controlled composition. Figure 5 selectively illustrates the ability of the best Detectron2 model under the increased complexity contained in these images.

### D. Images in the wild - multi-instance multi-plant classification with a single classifier

This category is by far the most difficult case we have attempted to solve. In this experiment, we sourced test images directly from field videos collected in the forests of Central Bali by our data collection team. In each case, the sample videos contain multiple instances of at least two distinct plant species represented as categories in our training sets. (See figure 6.)

### E. Evaluation of the experiments

None of the Detectron2 based single classifier approaches that performed well on single plant and composite dual-plant images were able to detect examples of multiple plant species in the wild. We trained the Detectron2 models (pretrained on the COCO dataset) for epochs of 3000 and 4000 with a learning rate of 0.00025, limited training to 2 images per batch and setting the region of interest box generation managing parameters to 128 per image. Both training and evaluation confidence thresholds were set to 50%. In order to evaluate each model's performance, we calculate the *recall*, *precision* and *f-measure* for each test image, as follows. For each test image, let *t* denote the actual set of species contained in the image and let *p* denote the set of species predicted by the model.

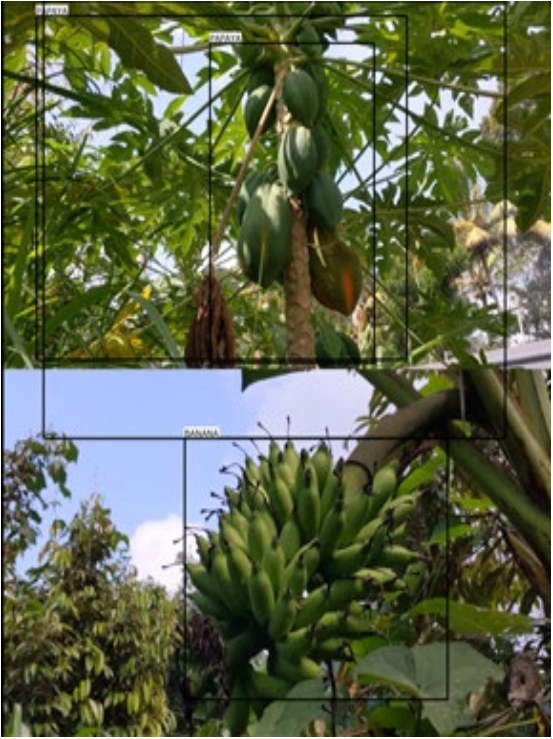

Fig. 5. Synthetic images (two species per composite image). Detectron2 D2.D results on the synthetic data set. Top: classifier correctly detects papaya fruit on a tree. Bottom: classifier detects a banana branch.

The recall, precision, and f-measure values for this prediction are defined as:

$$recall = (|t \cap p|)/(|t|) \quad precision = (|t \cap p|)/(|p|)$$

$$fmeasure = 2/(1/recall + 1/precision)$$

TABLE 3. Evaluation of Detectron2 classification variants on three datasets: bali-26 samples (single species), synthetic images (composed of two different species from bali-26), mixed plants in the wild (containing two species represented in the bali-26 collection)

| Detectron2 | data set | recall | precision | f-measure |
|---|---|---|---|---|
| D2.A | synthetic image set | 0.35 | 0.34 | 0.32 |
| D2.A | images in the wild | 0.41 | 0.31 | 0.32 |
| D2.A | bali-26 | 0.67 | 0.37 | 0.43 |
| D2.B | synthetic image set | 0.28 | 0.21 | 0.22 |
| D2.B | images in the wild | 0.26 | 0.21 | 0.21 |
| D2.B | bali-26 | 0.56 | 0.29 | 0.35 |
| D2.C | synthetic image set | 0.45 | 0.47 | 0.43 |
| D2.C | images in the wild | 0.26 | 0.43 | 0.32 |
| D2.C | bali-26 | 0.59 | 0.30 | 0.37 |
| D2.D | synthetic image set | 0.36 | 0.37 | 0.34 |
| D2.D | images in the wild | 0.44 | 0.44 | 0.41 |
| D2.D | bali-26 | 0.57 | 0.30 | 0.36 |

The *recall* measures the fraction of actual species in the image detected by the model. The *precision* measures how many of the detected species are actually present in the image. The *f-measure* is the harmonic mean of *recall* and *precision*, and provides a unified measure of the model performance. If the model fails to detect any species in an image, i.e., $p = \{\}$, the precision is 0. Here, we report the average values for the three metrics across all test images. The models were evaluated on three datasets mentioned above, the single species **bali-26** dataset, the **synthetic image set** with two species manually collated into a single image, and the **images in the wild** dataset with images collected in the field with at least two distinct species. All images were extracted from video. Overall, we observe that the different models have varying relative performance for the three different datasets. In particular, the D2.A model achieves solid (69% @k=1 after 3000 epochs), if not spectacular, results on the single plant instance training set. Even with a comparatively small fine-tuning training set, our best classifier candidate was able to detect and localize multiple instances of single plants as well as plant elements (stems and leaves).

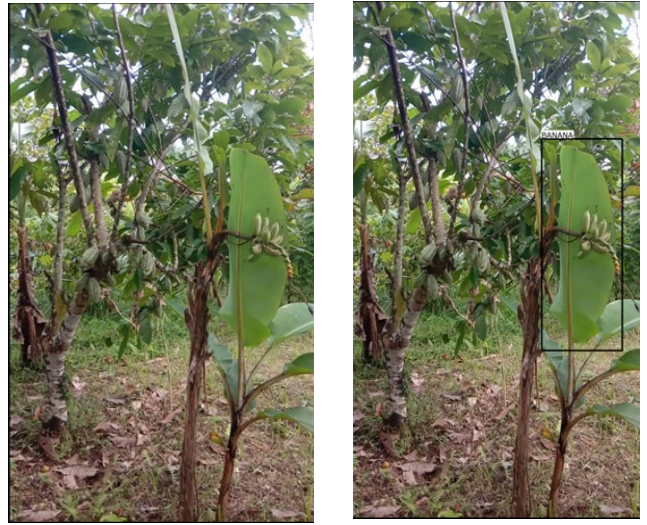

Fig. 6. Images in the wild. Left: Detectron2 D2.B on the images in the wild data set. It detects neither banana nor cacao Right. Detectron2 D2.C results on images in the wild. It detects one instance of banana but cannot detect the cacao plant and fruit.

V. IMAGE CLASSIFICATION WITH THE HELP OF VIDEO

As discussed above, we have not been able to identify a CNN system that can reliably detect multiple different plant species from video documentation in the wild in single images. While we continue to search for the 'perfect' classifier, we have more recently shifted attention to approaches that are less reliant on the performance of one classifier in isolation. To that end, we have created a novel classification scheme that takes properties of streaming video and inadequate individual classifiers into account.

*A. Making use of video capture and structure*

The provenance of a machine learning dataset can provide information to a classification algorithm that the data itself does not readily disclose. For example, recent research has made use of structural properties of satellite imagery (i.e., constant perspective and distance, significance of local-level image structure) to reduce the computational complexity of a machine learning classification approach [11]. Here, we focus on the structural properties of streaming video and the use of video by human videographers to improve machine learning.

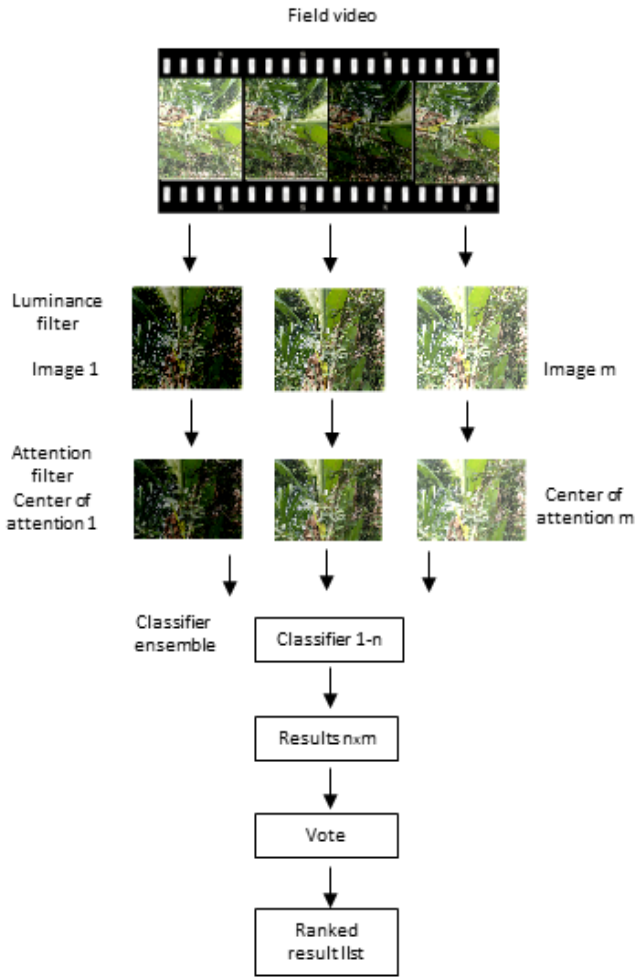

Fig. 7. Video enhanced image classification pipeline

Video streams contain more information than individual images. They contain time series information across image frames, and they contain variations in lighting and camera motion, for example. When a short video stream focuses on only one scene in a shot of several seconds, the time, lighting, and attention focal information can be used to create a series of images that are semantically identical, i.e. depicting the exact same plants in the wild, across slightly varying angles and lighting conditions. Our approach makes use of temporal context and adds voluntary and involuntary (indirect) cues contained within the structure of video. Voluntary cues are information that the image producer purposely inscribes into the image as he/she collects the data. In the test example, the voluntary information is limited to the *area of attention*, the central focus of the video frame. The involuntary (indirect) information is information that is casually collected as a side effect of the data collection effort and is typically not used as part of the classification task. The test case focuses on the variability of lighting conditions as one type of involuntary (indirect) information.

While traffic cameras and camera traps also produce image streams with temporal context that enables the system to determine if previous images are relevant to a current image [1], camera traps remain immobile. Consequently, they can rely on high precision spatial continuity, but also cannot make use of information generated by changing camera pose and position. Combining temporal, voluntary and indirect cues produces a set of varied image sections of constant semantics. This combination of a variety of image details and constancy in image content can facilitate the work of a classifier by offering more variation to a single labeled theme.

*B. An ensemble of imperfect classifiers*

Since no single classifier seems to be able to handle the complexity of our images, we have experimented with combining results from multiple classifiers to an *ensemble*. In our test case, the classifier ensemble consists of the four Faster R-CNN classifier architectures listed in Table 2, all fine-tuned on 100 (94) sets of 26 images from the bali-26 dataset. Each of these classifiers detects a unique set of probable results, and each of the classifiers responds differently to the slight luminescence variations in the sample images collected from the video stream, producing correct answers in addition to false positives and false negatives.

*C. Creating consensus across multiple imperfect classifiers*

We cannot know a priori which results of the individual classifiers are correct. We address this problem by combining the outputs of all classifiers to increase confidence in the cumulative results. Key to this approach is the assumption that results shared across classifiers are more likely to be correct. Results that occur in only one classifier, on the other hand, are considered more likely to be spurious. We tally all the responses (across all test images and classifier architectures) and rank the results by frequency of detected category (at a specified probability threshold). From this ranked list, we pick the top items as the most likely results. Below is an example of the output of this process on a 19-second field video depicting cacao and banana in a forest full of rich Balinese plant life.

*TALLY: {'sugarpalm': 6, 'cacao': 4, 'taro': 1, 'banana': 3, 'bamboo': 1, 'dragonfruit': 1}*

*FINAL TALLY above single instance ['sugarpalm', 'cacao', 'banana']*

The system correctly detects the presence of cacao and banana, but falsely registers that sugarpalm is dominantly represented (it is not). Compare this more complete result with that of our best single classifier outputs (figure 6) that detected at most one plant (banana). A link to a Jupyter notebook with this classification pipeline is listed in the *supporting materials* below. While this example is intended to illustrate the feasibility of the proposed concept, it solely demonstrates a new way of making use of information inherent to streaming media. We believe that the approach can be made applicable to other image-from-video classification challenges, and that the approach can assist in multi-evidence machine learning classification in general.

## VI. Discussion

Computer science has a very specific conceptual framing of 'images in the wild'. Real images in the wild are far richer and more difficult to deal with computationally, as our comparison between entropy in the Imagenet and bali-26 datasets demonstrates. This under-representation of image complexity is significant not only because of its consequences for classification algorithm design, but also because of the way it suggests that one must 'order' nature to make it accessible to technical processing, a position established through the history of botany's plant preparation practices, and continuing with the development of plant monitoring algorithms fine-tuned to plants grown in highly controlled laboratory and factory settings. While far from perfect, our process allows us to contribute to a still inadequately established discourse on the 'social construction of reality' [4] produced in the creation of image training data in computer science.

While we have pointed out how botany practices of controlled plant specimen collection influence image data collection practices, as well as how the manner in which classifiers are trained on them impacts classifier performance, the (reductionist) influence of data collection on machine learning is not limited to the field of botany and ethnobotany. The cost saved in 'conveniently' scrapping image data from the internet, for example, is a cost incurred in compromised representations in classification machinery. As such, this project demonstrates the need for rich and more realistic data to better represent the richness of the real world to computer systems designed to reason on the world. Furthermore, this project points to a persistent blind spot in data provenance within machine learning, namely the fact that hardly any training datasets emanate from the Global South [15]. With a focus on the deliriously rich plant life of Southeast Asia, the bali-26 collection contributes to diversity in machine learning data sets, allowing researchers to test their algorithms on alluring and complex imagery of non-western origin, as well as allowing students to study images of southeast Asian plants instead of cars, trains and planes as they learn how to craft and critique machine learning algorithms. In its surprisingly rich variety, immersion into the world represented in bali-26 creates a venue through which one might reflect upon which types of A.I. activities are desirable, and which types of input - such as granular location data [14] can be avoided.

### Acknowledgments

This project is supported in part by a grant from Microsoft's *AI for Earth Initiative*.

### Supporting Materials

*https://github.com/realtechsupport/video-informed-image-classification*